\documentclass[letterpaper]{article} 
\usepackage[]{aaai23}

\usepackage{amsmath,amsfonts,bm}

\def\eqref#1{equation~\ref{#1}}

\def\1{\bm{1}}

\DeclareMathAlphabet{\mathsfit}{\encodingdefault}{\sfdefault}{m}{sl}
\SetMathAlphabet{\mathsfit}{bold}{\encodingdefault}{\sfdefault}{bx}{n}

\DeclareMathOperator*{\argmax}{arg\,max}

\usepackage{times}  
\usepackage{helvet}  
\usepackage{courier}  
\usepackage[hyphens]{url}  
\usepackage{graphicx} 
\usepackage{natbib}  
\usepackage{caption} 
\DeclareCaptionStyle{ruled}{labelfont=normalfont,labelsep=colon,strut=off} 
\usepackage{algorithm}
\usepackage{algorithmic}

\usepackage{color}
\usepackage{booktabs}

\newcommand{\dplshort}{DPL}
\newcommand{\dpllong}{Domain Prompt Learning}

\usepackage{newfloat}
\usepackage{listings}
\floatstyle{ruled}
\newfloat{listing}{tb}{lst}{}
\floatname{listing}{Listing}
\usepackage{hyperref}

\title{Domain Prompt Learning for Efficiently Adapting CLIP to Unseen Domains}
\author {
    Xin Zhang,\textsuperscript{\rm 1}
    Shixiang Shane Gu \textsuperscript{\rm 1,2}
    Yutaka Matsuo, \textsuperscript{\rm 1}
    Yusuke Iwasawa, \textsuperscript{\rm 1}
}
\affiliations {
    \textsuperscript{\rm 1} The University of Tokyo\\
    \textsuperscript{\rm 2} Google Research, Brain Team\\
}

\usepackage{bibentry}

\begin{document}

\maketitle

\begin{abstract}

Domain generalization (DG) is a difficult transfer learning problem aiming to learn a generalizable model for unseen domains. 
Recent foundation models (FMs) are robust to many distribution shifts and, therefore, should substantially improve the performance of DG.
In this work, we study generic ways to adopt CLIP, a Visual-Language Foundation Model, for DG problems in image classification.
While ERM greatly improves the accuracy with bigger backbones and training datasets using standard DG benchmarks, fine-tuning FMs is not practical in many real-world situations.
We propose \dplshort~(\dpllong) as a novel approach for domain inference in the form of conditional prompt generation.
\dplshort~achieved a significant accuracy improvement with only training a lightweight prompt generator (a three-layer MLP), whose parameter is of equivalent scale to the classification projector in the previous DG literature.
Combining \dplshort~with CLIP provides surprising performance, raising the accuracy of zero-shot CLIP from 73.7\% to 79.3\% on several standard datasets, namely PACS, VLCS, OfficeHome, and TerraIncognita.
We hope the simplicity and success of our approach lead to broader adoption and analysis of foundation models in the domain generalization field.
Our code is available at \url{https://github.com/shogi880/DPLCLIP}
\end{abstract}

\section{Introduction}

\begin{figure}[h]
\begin{center}
\includegraphics[width=.9\linewidth]{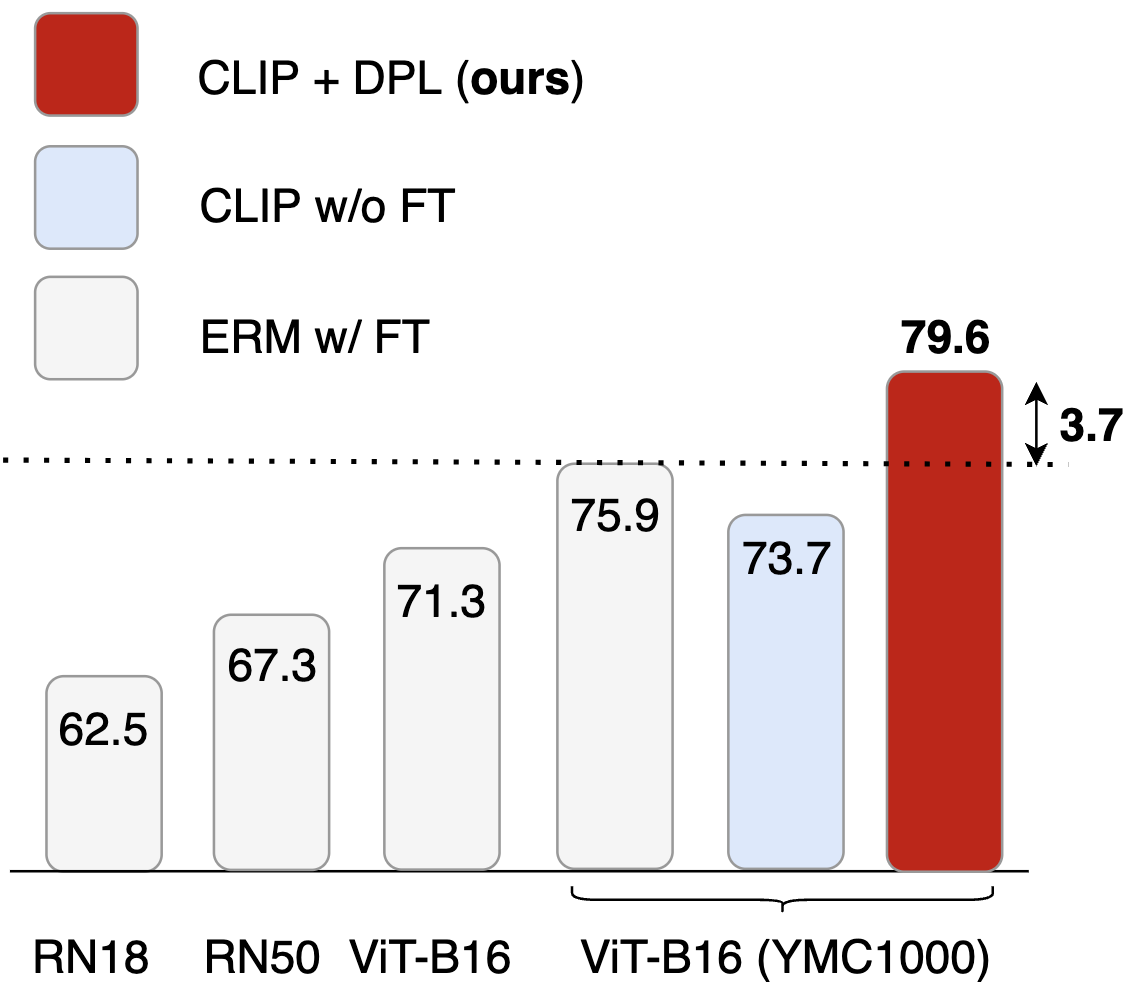}
\end{center}
  \caption{
  Bigger backbone (from ResNet18 to ViT-B16) and bigger pre-train dataset (from ImageNet to YMC1000) improve the performance of ERM on VLCS, PACS, OfficeHome, TerraIncognita.
  Even without fine-tuning the image encoder, our \dplshort~(\dpllong) effectively improves the performance of CLIP and outperforms the baseline ERM by a large margin (3.7\%).
  The CLIP w/o FT use the template prompt, such as `a photo of a \{class name\}'.
 }
\label{fig:first_image}
\end{figure}

Pre-training large vision models using web-scale images is an essential ingredient of recent success in computer vision. 
Fine-tuning pre-trained models, such as ResNet~\cite{he2015deep} and Vision Transformer (ViT)~\cite{dosovitskiy2020image} is the most popular paradigm for many downstream tasks. 
However, domain shifts pose a substantial challenge in real-world scenarios for successfully transferring models. 
Over the past decade, various studies on domain generalization (DG) have sought a systematic way to narrow the gap between source and target domains~\cite{zhou2021domain,wang2021generalizing,shen2021towards} aiming to build a model that generalizes to unseen domains. 
Despite the significant work on this front, machine learning systems are still vulnerable to domain shifts even after using DG methods \cite{gulrajani2020search}. 

Large pre-trained vision-language models like Contrastive Language-Image Pre-Training (CLIP) are an emerging category of models showing great potential in learning transferable representation across many vision tasks. 
At the core of CLIP is to learn image representations by contrasting them with the representations of text description of the image, such as `a photo of a \{class name\}'. 
The text description is often called \textit{prompt}, and its design is vital in enhancing CLIP performance. 
Notably, CLIP can handle unseen classes without fine-tuning them by adequately changing the text description using the target class name. 

\begin{figure*}[t]
\begin{center}
\includegraphics[width=1.\linewidth]{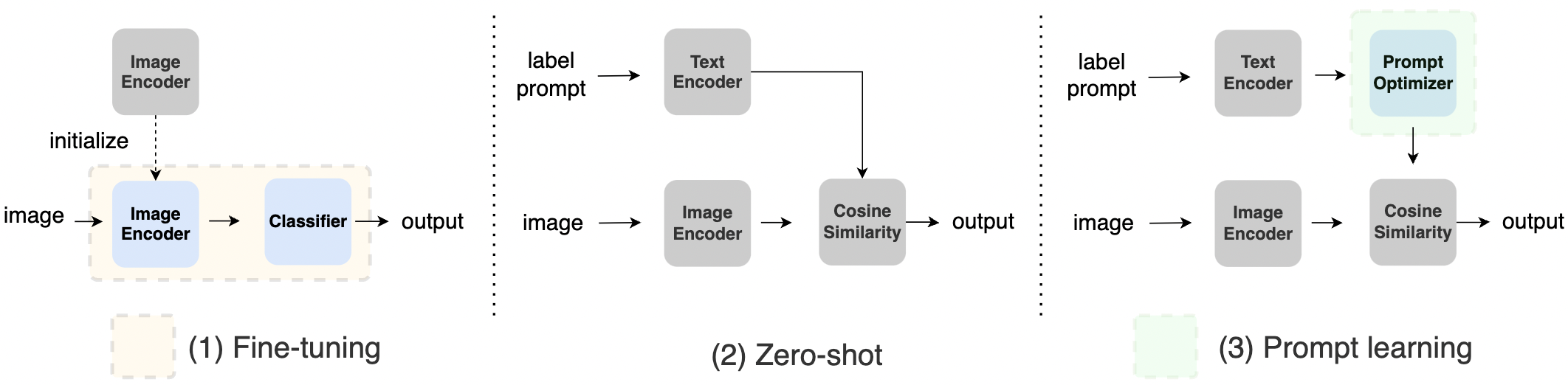}
\end{center}
  \caption{
  The concept illustration of three approaches to apply CLIP in DG.
  (1) Fine-tuning updates the CLIP's image encoder with a trainable classifier.
  (2) Zero-shot CLIP contrastive prediction with hand-craft prompts at the test time without updating parameters on the train domains.
  (3) Prompt learning trains a prompt optimizer then utilize the optimized prompts to prediction.
  Our \dplshort~is categorized to (3) Prompt learning, which trains a prompt generator in train phase and infers unseen domain to generate a domain-specific prompt.
 }
\label{fig:concept}
\end{figure*}

This paper investigates the robustness of CLIP against various distribution shifts using DomainBed~\cite{gulrajani2020search}, a recently proposed benchmark for DG setup. 
While prior works test various DG methods in the benchmark, the most studied only focused on medium-scale pre-trained models, such as ResNet18 or ResNet50. 
There are two na\"{i}ve approaches to leveraging CLIP in the DG setup~\autoref{fig:concept}. 
The first approach is fine-tuning the image encoder trained by CLIP, similar to the other vision models such as ResNet and ViT. 
We show that the backbone networks trained by CLIP substantially outperform many backbone networks trained solely on images, such as ResNet, big transfer \cite{kolesnikov2020big}, and vision transformer \cite{dosovitskiy2020image}. 
At the same time, however, fine-tuning sometimes degraded the performance on some domains, suggesting that fine-tuning possibly distorts good properties of pre-trained features~\cite{kumar2022fine}. 
Another na\"{i}ve approach is designing the template prompt, such as `a photo of a \{class name\}'. 
The clear merit of this approach is that it does not require optimizing any network and, therefore, keeps the representations learned via pre-training.
Despite its simplicity, we show that zero-shot CLIP is still more robust on many DG benchmarks than the vision backbones (e.g., ResNet18, ResNet50, ViT-B16) fine-tuned on source domains, while it is inferior to fine-tuning vision backbone trained by CLIP.

Based on the observations, we propose \dpllong~(\dplshort), a simple yet effective extension of CLIP in the DG setup.
A natural way to adapt the model is to add domain-specific features to the prompt template.
However, manually designing a prompt template is challenging in many cases due to its ambiguity. 
Instead, we propose \dplshort~for automatically generating a prompt that estimates domain-specific features given unlabeled examples from each distribution.
More specifically, \dplshort~trains a lightweight prompt generator using source domains, which outputs fixed-length continuous domain prompts given input images of each distribution while freezing other networks. 
During test-time, the prompt generator generates domain prompt given input images from the target distribution and adds them to the label prompts. 
Since the entire networks are frozen, the core properties of the pre-training would remain in DPL and are expected to improve CLIP performance in DG stably, as shown in our experiments. 

It is worth noting our work is not the first attempt to tune the prompt of CLIP. 
For example, \cite{gao2021clipadapter, zhou2021coop} have proposed optimizing continuous prompts on the target datasets, effectively improving CLIP performance. 
CoCoOp~\cite{zhou2022conditional}, as a contemporary work, trains a meta-net to generate a meta token for adapting to each instance. 
CoCoOp focuses on unseen classes and demonstrates its performance by transferring from ImageNet to the four specially designed ImageNet variants. 
This work focuses on the robustness of CLIP against distribution shifts, and proposes a generic way to extract a domain-specific features and improve performance on the target domain at test-time. 

We conduct experiments on four standard datasets included in DomainBed to evaluate \dplshort, following the experiment setup in\cite{gulrajani2020search,iwasawa2021testtime}, such as parameter tuning and model selection.
We show that CLIP with \dplshort~outperforms the strong baselines by a large margin, raising the accuracy from 73.7\% to 79.6\%~(\autoref{table:main_dg}).
Moreover, since \dplshort~can be seen as a kind of Test-Time Adaptation (TTA) method, we compare it with a series of SoTA TTA methods and demonstrate the efficiency of~\dplshort~(\autoref{table:main_tta}).
And lastly, through various ablation studies, we surprisingly found that frozen backbone outperforms fine-tuning on OfficeHome datasets for all of ResNet, DeiT~\cite{touvron2021training}, HViT, and ViT-B16~(\autoref{table:ablation_frezon}).
These results prove that \dplshort~is effective, and more importantly, they provide many insights for future works that apply CLIP on DG.

In summary, our main contributions are:
\begin{enumerate}
    \item We introduce CLIP to standard DG benchmark DomainBed via prompt learning.
    \item We propose \dpllong~(\dplshort), a novel approach of domain inference, to effectively help domain generalization by utilizing domain-specific features.
    \item We demonstrate the impressive empirical performance of \dplshort~by comparing with strong DG baselines and a series of state-of-the-art (SoTA) TTA methods.
\end{enumerate}

\section{Related Work}
\subsection{Domain Generalization}

Over the past decade, various approaches have been proposed to solve DG. 
Most prior works have focused on regularizing the model using the knowledge from multiple source domains.
For example, domain-invariant representation learning~\cite{ganin2016domainadversarial} is a major branch of domain generalization, aiming to reduce the domain gaps in the space of latent representations. 
There are many different approaches to measures the domain gaps, 
including adversarial classifier~\cite{li2018domain,ganin2015unsupervised,ganin2016domainadversarial},
kernel mapping~\cite{blanchard2011generalizing,grubinger2015domain}, 
metric learning~\cite{motiian2017unified,jin2020feature}, 
and invariant risk minimization~\cite{arjovsky2020invariant}. 
Similarly, several researchers have sought to generate samples with diverse styles so that models can learn domain-invariant features through them~\cite{shankar2018generalizing,zhou2020deep,borlino2021rethinking}. 
Other methods use meta learning to learn how to regularize the model to improve robustness~\cite{dou2019domain,li2017learning}.

Our work investigates the importance of the CLIP~\cite{radford2021learning} in DG, and proposes a lightweight way to adapt the CLIP for unseen domains. 
There are several recent observations to motivate us to benchmark CLIP in the DG setup. 
First, \cite{gulrajani2020search} shows that many prior approaches do not provide significant improvement compared to simple supervised learning. 
The results imply that regularizing the model is not sufficient to achieve high performance in DG. 
Secondly, despite significant related works, most studies have focused on medium-scale pre-trained models, such as ResNet18 or ResNet50, although very large-scale models often lead to substantial improvements.
Notably, the latest work~\cite{iwasawa2021testtime} compares more large-scale backbone networks, including big transfer \cite{kolesnikov2020big} (BiT-MR50x3, BiT-M-R101x3, and BiT-M-R152x4), vision transformer (ViTB16 and ViT-L16~\cite{dosovitskiy2020image}, Hybrid ViT, DeiT~\cite{touvron2021training}), and MLP-Mixer~\cite{tolstikhin2021mlpmixer} (Mixer-L16), and shows that the selection of backbone networks is important in DG.
In contrast with~\cite{iwasawa2021testtime}, we herein demonstrate that CLIP performs surprisingly well without fine-tuning the entire model in source domains, which is time-consuming in practice.

From the methodological perspective, our work relates to several prior works that have attempted leveraging domain features rather than discarding them~\cite{ganin2016domainadversarial,zhou2020deep,borlino2021rethinking}. 
While these works focused on the standard vision backbone, we propose a CLIP-specific approach to leverage the domain features by combining these features with prompt tuning. 

\subsection{Test Time Adaptation}
\label{sec:tta}
Regarding the problem setup, our work can also be seen as Test-Time Adaptation (TTA).
The concept of TTA is updating a part of networks to minimize the prediction entropy for adapting the model to an unseen domain robustly at the test time.
Pseudo Label~\cite{lee2013pseudo} updates entire networks and Tent~\cite{wang2020tent} updates the BN parameters.
SHOT\cite{liang2020we} update feature extractor and minimizes a diversity regularizer and pseudo-label loss, not only prediction entropy.
Instead of minimizing prediction entropy at the test time, we infer domain information and generate a domain-specific prompt to adapt CLIP to an unseen target domain.

Our work also relates to~\cite{iwasawa2021testtime} in that both approaches modulate their prediction given the unlabeled data available at test time. 
Specifically, \cite{iwasawa2021testtime} proposes T3A that replaces the linear classifier using pseudo-labeling and prototypical classification and shows that it stably improves the performance in unseen domains. 
However, T3A cannot be directly applied to CLIP, as it assumes a simple linear classifier that CLIP does not employ.

\subsection{Prompt Learning}
\label{sec:prompt}
The success of GPT-3 demonstrated the importance of prompt tuning. 
There are various prompting strategies, such as discrete natural language prompts and continuous prompts~\cite{liu2021pretrain}.
PADA~\cite{bendavid2021pada} proposed a domain adaptation algorithm that trains T5~\cite{raffel2019exploring}, a language foundation model, to generate unique domain-relevant features for each input.
PADA uses discrete prompts for the NLP applications and differs from our \dplshort~with continuous prompts in computer vision.
On the other hand, many recent works~\cite{li2021prefix,lester2021power} directly tuning prompts in continuous vector forms, and P-Tuning v2~\cite{liu2021ptuning} showed that continuous prompt tuning achieves the same performance as fine-tuning in various settings.

Because of the successful applications of CLIP, prompt tuning is also of great interest in computer vision. 
Context Optimization (CoOp~\cite{zhou2021coop}) demonstrated that CLIP performance is susceptible to prompts and that a suitable prompt can improve performance for the image recognition task.
CLIP-Adapter~\cite{gao2021clipadapter} was proposed to learn with an additional adapter network.
\cite{ge2022domain} adapts CLIP using contrastive learning in the Unsupervised Domain Adaption setup.
Unlike these works, which need to access the image or class labels in the target domain, 
we adapt CLIP to an unseen domain with a generated domain prompt inferred from input images. 

\section{Method}

In this section, we first introduce the notations and definitions of DG following~\cite{wang2021generalizing}.
Then, we explain how to use CLIP in DG and introduce \dpllong~to enhance CLIP performance in DG.

\subsection{Problem Setup of DG}

\begin{figure}[h]
\begin{center}
\includegraphics[width=1.\linewidth]{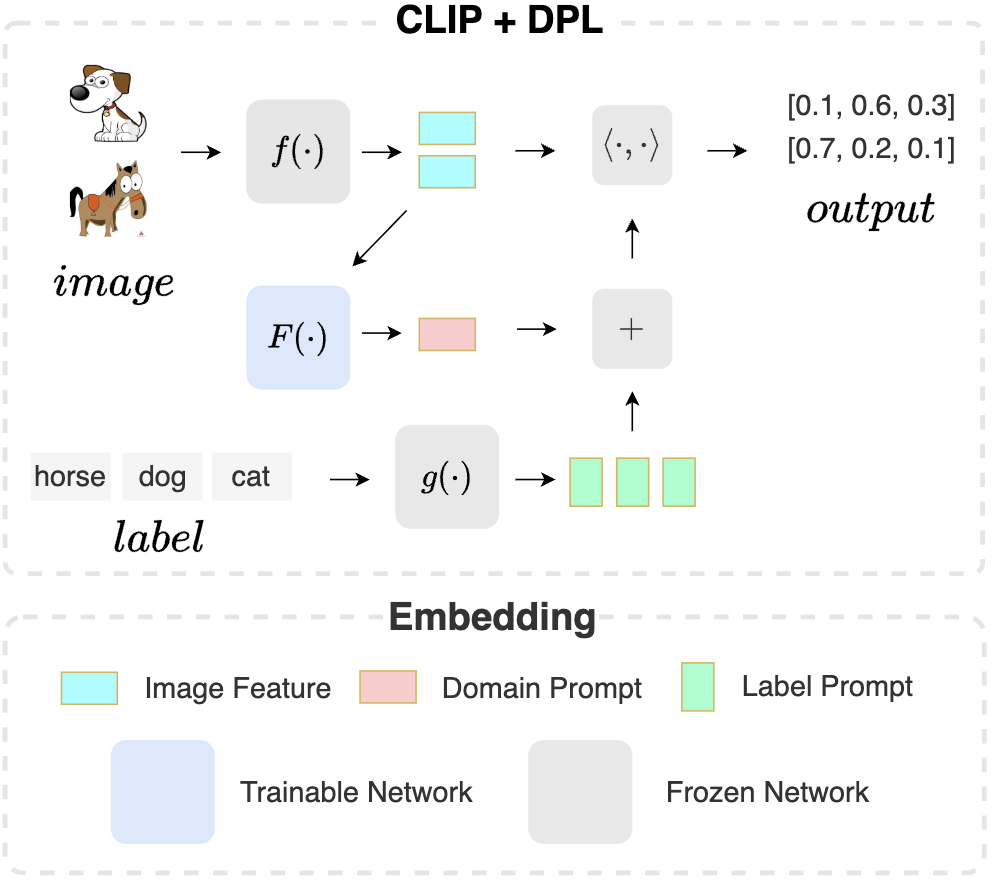}
\end{center}
  \caption{
  The architecture of CLIP + \dplshort.
  The only one network we trained is the prompt generator $F(\cdot)$, which is colored in blue.
  First, the input images are encoded to obtain image features with the frozen CLIP’s image encoder $f(\cdot)$.
  The image features are fed into the domain prompt generator $F(\cdot)$ to generate a domain prompt. 
  Simultaneously, all of labels are encoded using the frozen CLIP’s text encoder $g(\cdot)$ to obtain the label prompt embeddings.
  Secondly, the domain prompt embeddings are added to the label prompt embeddings for calculating the similarity.
  Finally, to obtain the prediction output in probability, the cosine similarity $\langle\cdot,\cdot \rangle$ are calculated with image embeddings and domain prompt embeddings.
 }
\label{fig:architecture}
\end{figure}


Let $\mathcal{X}$ denote an input space and $\mathcal{Y}$ an output space.
A domain is composed of data that has been sampled from a distribution.
We denote the datasets from distribution as 
$\mathcal{S}^{i} = {(\bm{x}_{j}^{i}, y_{j}^{i})}^{n_i}_{j=1} \sim \mathcal{P}_{XY}^{i}$,
where 
$\bm{x} \in \mathcal{X} \subset \mathbb{R}^{d}$ is an input image, 
$\bm{y} \in \mathcal{Y}$ denotes the class associated with $\bm{x}$, and 
$\mathcal{P}_{XY}^{i}$ 
denotes the joint distribution of the sample and output label in the domain $i$.
$X, Y$ denote the corresponding random variables.

In DG, we are interested in predictor $h$ performance on data from an unseen domain $\mathcal{P}_{XY}^{t} \neq \mathcal{P}_{XY}^{i}$ for all $i$. 
Prior works fine-tuned a pre-trained image encoder $f$ (usually ResNet18 or ResNet50) in conjunction with a randomly initialized classification head $g$ (linear classifier), using data from multiple different datasets to achieve the goal. 
Specifically, given $M$ datasets $\mathcal{S}^{i}$ collected from various domains $i \in \{ 1, \cdots, M\}$,  $f$ and $g$ are updated by  
\begin{equation}
\label{eq:domain-generalization}
    \min_{f, g} \frac{1}{M}\sum_{i=1}^{M}\frac{1}{n_i}\sum_{j=1}^{n_i}\ell(g \circ f(\bm{x_j^i}), y_j^i),
\end{equation}
where $\ell(\cdot)$ is a loss function. 
In the simplest case, $\ell$ is a simple cross-entropy loss, and minimizing eq. \ref{eq:domain-generalization} is called empirical risk minimization (ERM).
As discussed in Section \autoref{subsec:dtl}, different methods in DG use other loss functions by designing regularization terms to prevent overfitting specific domains.
These datasets are frequently referred to as source domains, and they are distinguished as target domains where we want the model to perform well.

\subsection{Na\"{i}ve Approaches for Using CLIP in DG}
CLIP consists of two parts: an image encoder $f_{clip}$ and a language model $g_{clip}$. 
CLIP classifies the image features based on the similarity between embedding of a text prompt $\bm{p}$, such as `dog' or `a photo of a {class label},' rather than initially using the classification head trained from scratch. 
Specifically, given an image $\bm{x}$ and $K$ class prompt $\bm{p}_k$, CLIP output a prediction using both $f_{clip}$ and $g_{clip}$:
\begin{equation}
\label{eq:domain-generalization}
    \hat{y}_{clip} = \argmax_{k}\langle f_{clip}(\bm{x}), g_{clip}(\bm{p}_k)\rangle
\end{equation}
where $K$ is the number of categories and $\langle \cdot,\cdot \rangle$ is cosine similarity.

To demonstrate how powerful the representation of massively pre-trained models (CLIP) for DG setup, we tested following two na\"{i}ve approaches to use CLIP in DG setups: fine-tuning and zero-shot.
Firstly, we evaluated CLIP in a zero-shot manner; i.e., we freeze both the image encoder and the language model, and substitute the class labels used in each dataset for the text prompt $\bm{p}$. 

Secondly, we can use the image encoder $f_{clip}$ as an alternative to the standard image backbones, such as ResNet and ViT. 
In this setup, we train $f_{clip}$ by using the datasets $\mathcal{S}^{i}$ from multiple source domains $i$, similar to the standard DG setup. 
We can use any algorithms tailored for DG setup, such as DANN and CORAL during fine-tuning. 
While it is powerful as shown in experiments, it requires additional computational costs to re-train such large models entirely. 
Besides, good properties of massive pre-training might be distorted during fine-tuning, as highlighted by the performance degradation compared to zero-shot approach. 

In summary, zero-shot approach is computationally effective yet less expressive, and fine-tuning can leverage the knowledge of source datasets but it is computationally heavy and possibly distort good representations learned during pre-training. 
Based on the observation, we propose a novel approach to design the prompt $p$ to improve the performance in an unseen domain without fine-tuning the entire model. 

\subsection{\dpllong~for CLIP in DG}

\begin{table*}[h]
\begin{center}
\begin{tabular}{lc|cccc|l}
\toprule
\textbf{DomainBed} & category & \textbf{VLCS} & \textbf{PACS} & \textbf{OfficeHome} & \textbf{Terra} & \textbf{Avg} \\
\midrule
ERM (CLIP) & Fine-tuning & 82.7 $\pm$ 0.3 & 92.9 $\pm$ 1.9 & 78.1 $\pm$ 2.1 & 50.2 $\pm$ 1.7 & 75.9\\
CORAL & Fine-tuning & 82.0 $\pm$ 0.2 & 93.2 $\pm$ 1.1 & 78.9 $\pm$ 1.9 & \textbf{53.5 $\pm$ 0.7} & 76.9 \\
DANN & Fine-tuning & \underline{83.2 $\pm$ 1.2} & 93.8 $\pm$ 1.3 & 78.8 $\pm$ 1.1 & 52.2 $\pm$ 2.0 & \underline{77.0} \\
\midrule
CLIP & Zero-shot & 76.6 $\pm$ 0.0 & 95.8 $\pm$ 0.1 & 79.9 $\pm$ 0.1 & 36.4 $\pm$ 0.1 & 72.2  \\
CLIP (template prompt)& Zero-shot & 82.3 $\pm$ 0.1 & 96.1 $\pm$ 0.1 & \underline{82.3 $\pm$ 0.2} & 34.1 $\pm$ 0.1 & 73.7 \\
\textbf{CLIP + \dplshort~(ours)}& Prompt learning & \textbf{84.3 $\pm$ 0.4} & \textbf{97.3 $\pm$ 0.2} & \textbf{84.2 $\pm$ 0.2} & \underline{52.6 $\pm$ 0.6} & \textbf{79.6} \\
\midrule
\end{tabular}
\caption{
Comparison experiments on VLCS, PACS, OfficeHome, and TerraIncognita. 
The best results are in bold, and the second-best results are underlined.
CLIP (standard template) indicates using `a photo of a \{class name\}' prompt.
Following the experiment setup in~\cite{iwasawa2021testtime}, reported results are the mean and std with seed=\{1, 2, 3\}.
}
\label{table:main_dg}
\end{center}
\end{table*}

As discussed in Section \ref{sec:prompt}, designing a prompt is a powerful approach to improve the performance of the transformer-based models. 
It is powerful and should also be easier to train because the dimension of prompts is significantly smaller than the entire parameters of $f$ and $g$. 
For example, supposing we can access a supervised dataset from the target domain, we can optimize a prefix vector $\bm{p}_{pre}$ by simple supervised loss: 
\begin{equation}
\min_{\bm{p}_{pre}} \mathbb{E}_{x, y \sim \mathcal{S}} \ell(\hat{y}_{clip*}, y), 
\end{equation}
where $\hat{y}_{clip*}$ is 
\begin{equation}
    \hat{y}_{clip*} = \argmax_{k}\langle f_{clip}(\bm{x}), g_{clip}(\bm{p}_k^{*})\rangle,
\end{equation}
where $\bm{p}_k^{*}$ is a concatenation of trainable parameters $\bm{p}_{pre}$ and $\bm{p}_{k}$. 
Particularly, $g_{clip}$ outputs the fixed length vector regardless of the input dimension (i.e., size of $\bm{p}_{k}$). 
The size of $\bm{p}_{k}$ is a hyperparameter.

Unfortunately, this labeled training data for the target domain is unavailable in DG. 
Instead, we proposed \dplshort~to replace the optimization process of $\bm{p}_{pre}$ in each domain by training novel prompt generators $F(\cdot)$ that generate a prompt $\bm{p}_{pre}$ given small unlabeled images from a distribution. 
Specifically, we use a fully connected network $F(\cdot)$ to generate a prompt $p$ from input images: 
\begin{equation}
    \bm{p}_{ap}^{i} = \frac{1}{N} \sum_{j = 1}^{N}F(f(x_{j}^{i})), 
\end{equation}
where $N$ is the batch size for each domain, and $\bm{x}_{j}^{i}$ denotes the images from the $i$-th distribution. 
Given a batch of data from multiple source distributions, we use the following loss function to optimize $F$:
\begin{equation}
\min_{F} \frac{1}{M}\sum_{i=1}^{M}\frac{1}{n_i}\sum_{j=1}^{n_i}\ell(\hat{y}_{ap}^i, y_j^i),
\end{equation}
and 
\begin{equation}
    \hat{y}_{ap}^i = \argmax_{k}\langle f_{clip}(\bm{x^i}), g_{clip}(\bm{p}_k^{i})\rangle,
\end{equation}
where  $\bm{p}_k^{i}$ is a concatenation of pre-defined $\bm{p}_k$ and $\bm{p}_{ap}^i$ . 
The architecture of CLIP + \dplshort~is depicted in~\autoref{fig:architecture}.

\begin{table*}[t]
\begin{center}
\begin{tabular}{l|cccc|l}
\toprule
\textbf{Methods} & \textbf{VLCS} & \textbf{PACS} & \textbf{OfficeHome} & \textbf{Terra} & \textbf{Avg} \\
\midrule
ERM & 81.4 $\pm$ 0.3 & 91.9 $\pm$ 0.7 & 78.4 $\pm$ 1.1 & 47.8 $\pm$ 3.1 & 74.9 \\
+T3A & \textbf{82.2 $\pm$ 0.1} & 88.2 $\pm$ 0.0 & 76.9 $\pm$ 0.9 & 48.2 $\pm$ 3.2 & 73.9 \\
+Pseudo Label   & 81.1 $\pm$ 1.0 & 92.1 $\pm$ 0.4 & 78.0 $\pm$ 2.5 & \textbf{50.5 $\pm$ 4.4} & \underline{75.4} \\
+Pseudo Label$\star$ & \underline{82.1 $\pm$ 0.4} & 87.6 $\pm$ 0.0 & 76.6 $\pm$ 1.9 & 46.9 $\pm$ 3.1 & 73.3 \\
+Tent$\star$ & \textbf{82.2 $\pm$ 0.4} & 87.8 $\pm$ 0.0 & 76.5 $\pm$ 1.2 & 46.7 $\pm$ 3.2 & 73.3 \\
+SHOT & 80.4 $\pm$ 0.4 & \underline{93.8 $\pm$ 0.8} & \underline{80.7 $\pm$ 0.8} & 40.5 $\pm$ 1.7 & 73.9 \\
+SHOT$\star\star$ & 80.3 $\pm$ 0.5 & 91.2 $\pm$ 0.0 & 79.4 $\pm$ 1.1 & 40.6 $\pm$ 1.8 & 72.9 \\
\midrule
CORAL & 81.2 $\pm$ 0.3 & 91.1 $\pm$ 1.9 & 78.7 $\pm$ 0.8 & 48.6 $\pm$ 2.9 & 74.9 \\
+T3A & 80.8 $\pm$ 0.5 & 91.2 $\pm$ 1.9 & 79.1 $\pm$ 0.9 & 49.0 $\pm$ 3.0 & 75.0 \\
+Pseudo Label  & 80.0 $\pm$ 1.4 & 93.1 $\pm$ 2.0 & 79.8 $\pm$ 1.2 & 44.5 $\pm$ 3.3 & 74.4 \\
+Pseudo Label$\star$ & 81.4 $\pm$ 0.1 & 91.2 $\pm$ 1.9 & 78.8 $\pm$ 0.9 & 48.6 $\pm$ 2.9 & 75.0           \\
+Tent$\star$ & 81.3 $\pm$ 0.2 & 91.2 $\pm$ 1.9 & 78.6 $\pm$ 0.8 & 48.7 $\pm$ 2.6 & 75.0           \\
+SHOT & 78.7 $\pm$ 1.9 & 93.0 $\pm$ 1.2 & \underline{80.7 $\pm$ 0.9} & 41.9 $\pm$ 2.0 & 73.6\\
+SHOT$\star\star$ & 78.5 $\pm$ 2.0 & 93.1 $\pm$ 1.1 & \underline{80.7 $\pm$ 0.9} & 41.9 $\pm$ 2.0 & 73.5 \\
\midrule
\textbf{CLIP + \dplshort~(ours)}& 81.0 $\pm$ 1.1 & \textbf{95.9 $\pm$ 0.0} & \textbf{82.3 $\pm$ 0.7} & \underline{49.4 $\pm$ 1.1} & \textbf{77.2}           \\
\bottomrule
\end{tabular}
\end{center}
\caption{
Comparison with TTA methods.
Here, $\star$ indicates updating the linear classifier, and $\star\star$ indicates updating the feature extractor to minimize entropy reported in table 3 of the T3A paper.
The best results are in bold, and second-best results are underlined.
All the experiments listed in this table are run on a cluster of A100 GPUs.
The numbers of ERM, CORAL, and CLIP + \dplshort~are different from \autoref{table:main_dg} because of the use of half precision floating point on A100.
}
\label{table:main_tta}
\end{table*}

\section{Experiment}

In this section, we experimentally demonstrate the effectiveness of \dplshort.
First, we clarify the important DG settings, including the datasets, hyperparameters, model selection strategy, and other implementation details.
Second, we show CLIP + \dplshort~outperforms the strong DG baselines and several SoTA TTA methods on DomainBed benchmark.
Finally, our ablation experiments, including variants backbone comparison and different prompt strategies study, provide meaningful insights of applying CLIP + \dplshort~to DG.

\paragraph{Datasets}
Following~\cite{iwasawa2021testtime}, we selected four real-world datasets from DomainBed benchmark, including VLCS~\cite{fang2013unbiased}, PACS~\cite{li2017deeper}, OfficeHome~\cite{venkateswara2017deep}, TerraIncognita~\cite{beery2018recognition}.
More details are provided in Appendix A.

\paragraph{Hyperparameters and model selection.} 
We set up experiments on DomainBed~\footnote{https://github.com/facebookresearch/DomainBed}, and implemented \dplshort~based on CLIP\footnote{https://github.com/openai/CLIP}.
We strictly followed the basic selection criterion~\cite{gulrajani2020search} and selected the hyperparameters using standard training-domain validation.
First, we split the data of each domain into 80\% and 20\% for the training model and select hyperparameters.
Then, we ran 20 trials at random across a joint distribution of all hyperparameters. 
Next, we ran three trials of each hyperparameter setting, reserving one domain for testing and the rest for training. 
Finally, we selected the hyperparameters that maximize validation accuracy across the training domains and reported overall accuracy averaged across all three trials.

\paragraph{Detail of the implements}
As shown as in \autoref{fig:architecture}, we only trained a three-layer MLP as the domain prompt generator.
We used stochastic gradient descent~\cite{bottou2012stochastic} with momentum as an optimizer.
Refer to our source code for implement details.

\subsection{Comparison with existing DG methods}
\label{subsec:dg}

\paragraph{Baselines}
We compared our method to domain generalization algorithms, which fine-tune image features, and the handcrafted prompt for CLIP.
For DG, we trained CLIP image features (ViT-B16) using ERM, CORAL~\cite{sun2016deep}, and DANN~\cite{ganin2016domain}. 
Note that, as~\cite{gulrajani2020search} pointed out, ERM is a strong DG baseline when the experiments are fairly performed. 
For handcrafted prompt, we adopted three types prompt for CLIP including `\{class name\}', template prompt `a photo of a \{class name\}.', and \dpllong~`$v_{1}, v_{2}, ..., v_{n}$ \{class name\}.'.

All experiments listed in \autoref{table:main_dg} are based on the CLIP ViT-B16 backbone.
We observed that zero-shot CLIP could achieve an average of 72.2\% accuracy and an average of 73.7\% by using a template prompt.
Notably, \dplshort~improves CLIP performance to 79.6\% and outperforms all baselines, although ERM, CORAL, and DANN are fine-tuning their image encoder.
Based on this result, we infer that \dplshort~should be an effective method in DG.

Surprisingly, we found that fine-tuning the backbones hurts performance on PACS and OfficeHome.
We consider that fine-tuning causes the model to overfit in the source domain in the case where the pre-training domain is big enough to cover the target domain.
On the other hand, the models perform better with fine-tuning on Terra, with a high likelihood of being not covered by the CLIP's pre-training dataset.
It is worth noting that our \dplshort~can effectively trade-off well between both cases.

\subsection{Comparison with existing TTA methods}

\dplshort~is generated by extracting domain features from a batch of input images.
As discussed in \autoref{sec:tta}, \dplshort~can be considered as a TTA method.
Therefore, we performed a fair comparison with several TTA algorithms to validate \dplshort.

\paragraph{Baselines}

Following~\cite{iwasawa2021testtime}, we adopted the baselines including Pseudo Label, SHOT, Tent, and T3A with batch size equal to 64 during the test time.
We trained all models with the same CLIP ViT-B16 backbone.
All the experiments follow the model selection, hyperparameter selection strategy, and evaluation method proposed in T3A and DomainBed.

As shown in \autoref{table:main_tta}, \dplshort~beats the most effective TTA methods on four datasets.
The result demonstrates that \dplshort~can consistently improve the model's generalization performance at test time.
We believe this is sufficient evidence that the central concept of \dplshort, extracting unseen domain features to help model adapting at the test time, is practical.

\subsection{Backbone Ablation}
\paragraph{Different Backbones} 
Many proposed DG methods are evaluated using the standard ResNet backbones.
However, more and more large models are being studied, and their validity is being experimentally demonstrated~\cite{bommasani2021opportunities,wang2022deepnet}.
Therefore, we reported the performance of ResNet18 and ResNet50, Mixer-16~\cite{tolstikhin2021mlpmixer}, Vision Transformer (ViT)~\cite{dosovitskiy2020image} and several variations of ViT, such as BiT~\cite{kolesnikov2020big}, DeiT~\cite{touvron2021training}, HViT, and Mutual Information Regularization with Oracle (MIRO)~\cite{cha2022domain} in \autoref{table:ablation_backbone}.

As a result, we discovered that the CLIP ViT-B16 backbone trained on YFCC100M~\cite{thomee2016yfcc100m} performs as well as HViT.
Moreover, CLIP + \dplshort~surpassed most of the backbones, including HViT and MIRO.
Notably, \dplshort~only trains a three-layer MLP, in contrast to others fine-tuning their backbones.
We observed that the SoTA performance is provided by MIRO using RegNetY-16GF backbone with SWAG pre-training and combined with Stochastic Weight Averaging Densely (MIRO + SWAG~\cite{singh2022revisiting} + SWAD~\cite{cha2021swad}).
The simple \dplshort~can achieve close performance (difference of ~1.9\%).
Although comparing with different pre-training datasets and different parameters is unfair, this result demonstrates that \dpllong~can efficiently adapt CLIP to unseen domains.

\begin{table}[t]
\begin{center}
\resizebox{0.48\textwidth}{!}{
\begin{tabular}{l|cccc|l}
\toprule
\textbf{Backbone Model}  & \textbf{VLCS} & \textbf{PACS} & \textbf{OfficeHome}  & \textbf{Terra} & \textbf{Avg}      \\
\midrule
ResNet18$^\dag$  & 73.2 $\pm$ 0.9 & 80.3 $\pm$ 0.4 & 55.7 $\pm$ 0.2 & 40.7 $\pm$ 0.3 & 62.5\\
ResNet50$^\dag$  & 75.5 $\pm$ 0.1 & 83.9 $\pm$ 0.2 & 64.4 $\pm$ 0.2 & 45.4 $\pm$ 1.2 & 67.3\\
\midrule
Mixer-L16$^\dag$ & 76.4 $\pm$ 0.2 & 81.3 $\pm$ 1.0 & 69.4 $\pm$ 1.6 & 37.1 $\pm$ 0.4 & 66.1\\
\midrule
BiT-M-R50x3$^\dag$ & 76.7 $\pm$ 0.1 & 84.4 $\pm$ 1.2 & 69.2 $\pm$ 0.6 & 52.5 $\pm$ 0.3 & 70.7\\
BiT-M-R101x3$^\dag$  & 75.0 $\pm$ 0.6 & 84.0 $\pm$ 0.7 & 67.7 $\pm$ 0.5 & 47.8 $\pm$ 0.8 & 68.6\\
BiT-M-R152x2$^\dag$  & 76.7 $\pm$ 0.3 & 85.2 $\pm$ 0.1 & 71.3 $\pm$ 0.6 & 51.4 $\pm$ 0.6 & 71.1\\
ViT-B16$^\dag$ & 79.2 $\pm$ 0.3 & 85.7 $\pm$ 0.1 & 78.4 $\pm$ 0.3 & 41.8 $\pm$ 0.6 & 71.3\\
ViT-L16$^\dag$ & 78.2 $\pm$ 0.5 & 84.6 $\pm$ 0.5 & 78.0 $\pm$ 0.1 & 42.7 $\pm$ 1.9 & 70.9\\
DeiT$^\dag$    & 79.3 $\pm$ 0.4 & 87.8 $\pm$ 0.5 & 76.6 $\pm$ 0.3 & 50.0 $\pm$ 0.2 & 73.4  \\
HViT$^\dag$    & 79.2 $\pm$ 0.5 & 89.7 $\pm$ 0.4 & 80.0 $\pm$ 0.2 & 51.4 $\pm$ 0.9 & 75.1\\
\midrule
MIRO* & 79.0 $\pm$ 0.0 & 85.4 $\pm$ 0.4 & 70.5 $\pm$ 0.4 & 50.4 $\pm$ 1.1 & 71.3  \\
MIRO + SWAD*  & 79.6 $\pm$ 0.2 & 88.4 $\pm$ 0.1 & 72.4 $\pm$ 0.1 & 52.9 $\pm$ 0.2 & 73.3  \\
MIRO + SWAG* & 79.9 $\pm$ 0.6 & \textbf{97.4 $\pm$ 0.2} & 80.4 $\pm$ 0.2 & \underline{58.9 $\pm$ 1.3} & 79.2  \\
MIRO + SWAD + SWAG* & 81.7 $\pm$ 0.1 & 96.8 $\pm$ 0.2 & \underline{83.3 $\pm$ 0.1} & \textbf{64.3 $\pm$ 0.3} & \textbf{81.5}  \\
\midrule
\midrule
CLIP ViT-B16 & \underline{82.7 $\pm$ 0.3} & 92.9 $\pm$ 1.9 & 78.1 $\pm$ 2.1 & 50.2 $\pm$ 1.7 & 75.9\\
\textbf{CLIP+\dplshort}& \textbf{84.3 $\pm$ 0.4} & \underline{97.3 $\pm$ 0.2} & \textbf{84.2 $\pm$ 0.2} & 52.6 $\pm$ 0.6 & \underline{79.6} \\
\midrule
\end{tabular}}
\end{center}
\caption{
Results of ERM with various backbone networks on DG benchmark.
$^\dag$ indicates that the numbers are taken from Table 2 in~\cite{iwasawa2021testtime}.
* indicates the numbers are taken from MIRO~\cite{cha2022domain}.
The best scores are bolded, and the second-best scores are underlined.
}
\label{table:ablation_backbone}
\end{table}

\begin{table}[t]
\label{sample-table}
\begin{center}
\resizebox{0.48\textwidth}{!}{
\begin{tabular}{l|cccc|l}
\toprule
\textbf{Backbone} & \textbf{VLCS} & \textbf{PACS} & \textbf{OfficeHome} & \textbf{Terra} & \textbf{Avg} \\
\midrule
(1) Frozen & 76.0 $\pm$ 0.3 & 66.0 $\pm$ 0.7 & 61.7 $\pm$ 0.5 & 25.5 $\pm$ 1.8 & 57.3 \\
(2) ResNet18$^\dag$ & 73.2 $\pm$ 0.9  & 80.3 $\pm$ 0.4 & 55.7 $\pm$ 0.2 & 40.7 $\pm$ 0.3 & 62.5 \\
(2) - (1) & \textcolor{red}{-2.8} & +14.3 & \textcolor{red}{-6.0} & +15.2 & +4.2 \\
\midrule
(1) Frozen  & 77.4 $\pm$ 0.3 & 67.2 $\pm$ 0.4 & 68.0 $\pm$ 0.3 & 35.4 $\pm$ 1.5 & 62.0 \\
(2) ResNet50$^\dag$ & 75.5 $\pm$ 0.1  & 83.9 $\pm$ 0.2 & 64.4 $\pm$ 0.2 & 45.4 $\pm$ 1.2 & 67.3 \\
(2) - (1) & \textcolor{red}{-1.9} & +16.7& \textcolor{red}{-3.6} & +10.0 & +5.3 \\
\midrule
(1) Frozen  & 77.5 $\pm$ 0.4 & 74.3 $\pm$ 0.3 & 77.4 $\pm$ 0.2 & 43.4 $\pm$ 0.3 & 68.2 \\
(2) DeiT $^\dag$ & 79.3 $\pm$ 0.4 & 87.8 $\pm$ 0.5 & 76.6 $\pm$ 0.3 & 50.0 $\pm$ 0.2 & 73.4 \\
(2) - (1) & +1.8 & +13.5 & \textcolor{red}{-0.8} & +6.6 & +5.2 \\
\midrule
(1) Frozen & 79.2 $\pm$ 0.1 & 76.6 $\pm$ 0.4 & 81.1 $\pm$ 0.2 & 35.7 $\pm$ 0.7 & 68.1   \\
(2) HViT$^\dag$  & 79.2 $\pm$ 0.5 & 89.7 $\pm$ 0.4 & 80.0 $\pm$ 0.2 & 51.4 $\pm$ 0.9 & 75.1 \\
(2) - (1) & \textcolor{red}{-0.0} & +13.1 & \textcolor{red}{-1.1} & +15.7 & +7.0 \\
\midrule
(1) Frozen   & 82.6 $\pm$ 0.3 & 96.9 $\pm$ 0.1 & 83.2 $\pm$ 0.2 & 46.5 $\pm$ 2.1 & 77.3 \\
(2) CLIP ViT-B16 & 82.7 $\pm$ 0.3 & 92.9 $\pm$ 1.9 & 78.1 $\pm$ 2.1 & 50.2 $\pm$ 1.7 & 75.9 \\
(2) - (1) & +0.1 & \textcolor{red}{-4.0} & \textcolor{red}{-5.1} & +3.7 & \textcolor{red}{-1.4} \\
(3) \dplshort~(ours)& \textbf{84.3 $\pm$ 0.4} & \textbf{97.3 $\pm$ 0.2} & \textbf{84.2 $\pm$ 0.2} & \textbf{52.6 $\pm$ 0.6} & \textbf{79.6} \\
(3) - (1) & +1.7 & +0.4 & 1.0 & +6.1 & +2.3 \\
\midrule
\end{tabular}
}
\end{center}
\caption{
The results of frozen backbone ablation with ERM.
Each block represents a backbone.
Frozen means using the frozen backbone.
$^\dag$ indicates that the numbers are taken from Table 2 in~\cite{iwasawa2021testtime}.
The highlighted numbers indicate the Frozen ERM outperforms the standard ERM with fine-tuning backbones.
(3) refers to \dplshort, which scores beat all others and are bolded.
}
\label{table:ablation_frezon}
\end{table}

\paragraph{Frozen Backbone}
Fine-tuning a large model like CLIP or other Foundation Models necessitates much computing power.
\dplshort~also aims to adapt CLIP to the target domain with minimum computing.
We wondered if simply training an MLP classifier with the frozen backbone could aid model transfer and conducted the ablation experiments with five different backbones.

From \autoref{table:ablation_frezon}, we surprisingly found that Frozen ERM outperforms the standard ERM in OfficeHome with all the backbones. 
In VLCS, the performance of Frozen ERM is also unexpected.
These results show that fine-tuning hurts the model more than expected on specific datasets.
On the other hand, \dplshort~steadily improving the performance on all datasets demonstrates the robustness of \dplshort.

A similar phenomenon, fine-tuning does not constantly improve performance in DG, is also observed in \autoref{subsec:dg}.
Due to computing resource constraints, we only evaluated several backbones of varying sizes in this work.
We found several recent studies analyzing the same phenomenon, the effect of pre-training datasets and backbones on DG and Out-of-Distribution settings\cite{kim2022broad,wenzel2022assaying}.

\section{Conclusions}
We introduce CLIP to DG on DomainBed.
For this purpose, we proposed a novel approach called \dpllong~(\dplshort) for efficiently adapting CLIP to an unseen domain.
By generating the domain prompt conditional on input images, CLIP + \dplshort~brings substantial improvements over strong DG baselines and several effective TTA methods on DomainBed.
Then, we conducted ablation experiments with various backbones and Frozen ERM.
We verified that \dplshort~can stabilize performance and present meaningful insights about existing datasets and fine-tuning strategy of backbones.
We hope that our research will broaden and inspire the roles of prompt learning in domain transfer learning.

\subsection{Limitation}
\label{subsec:limitation}
\paragraph{Interpretability of Domain Prompt} To better perform, our \dplshort~is directly represented in a continuous vector form, which lacks interpretability. 
However, improving interpretability is an important research direction in both FM applications and Domain Generalization.
We consider producing discrete semantically informative prompts by some means is an exciting extension of \dplshort, even with some loss of precision.

\paragraph{Label Shift} From the technical perspective, \dplshort~cannot capture the domain shift outside of the images because \dplshort~uses domain features extracted from only images.
As a result, \dplshort~has no idea how to record such non-visual domain shift.
Unfortunately, the label shift exists in the actual-world applications~\cite{azizzadenesheli2019regularized}.
An innovative question is whether adding appropriate information to the Domain Prompt can help solve the label shift problem, such as a detailed textual description of the target domain.

\paragraph{Social impact perspective}
Many images and text descriptions of web data are used directly to train CLIP. 
Though CLIP benefits from low-cost data that do not require manual labeling, it inevitably includes a lot of bias and privacy in CLIP and other foundation models~\cite{bommasani2021opportunities}.
This requires us to spend more time paying attention to the opportunities and risks of Foundation Models.

\subsection{Future Work}
First and foremost, interpretability is critical in both Domain Transfer Learning and the Foundation Model. 
As discussed in \autoref{subsec:limitation}, \dplshort~introduce the possibility of using a large language model in DG in the form of prompt.
We will investigate this direction in our future work.

There are two simple and critical approaches to improving the performance of DG. 
One is to apply visual prompt tuning~\cite{jia2022visual} on the pure visual backbones, which can be used to more previous methods.
Another is focusing on a data-centric approach since we observe uneven data quality on the widely used datasets.

Finally, several recent studies systematically analyze the performance and shortcomings of large-scale pre-train models in the Out-of-Distribution generalization~\cite{cha2022domain,wenzel2022assaying}.
We hope that our results will inspire more research in this direction.

\bibliography{aaai23}

\newpage
\section{Appendix}
\subsection{Datasets}

Following~\cite{iwasawa2021testtime}, we selected four real-world datasets from DomainBed benchmark, including VLCS~\cite{fang2013unbiased}, PACS~\cite{li2017deeper}, OfficeHome~\cite{venkateswara2017deep}, TerraIncognita~\cite{beery2018recognition}.

VLCS~\cite{fang2013unbiased} gathers four photographic datasets 
$d \in$ \{Caltech101~\cite{fei2006one}, LabelMe~\cite{russell2008labelme}, SUN09~\cite{choi2010exploiting}, VOC2007~\cite{everingham2009pascal}\}, 
containing 10,729 samples of 5 classes.
PACS~\cite{li2017deeper}  comprises four domain datasets $d \in$ \{art, cartoons, photos, sketches\}, with 9,991 samples and 7 classes.
OfficeHome~\cite{venkateswara2017deep}  includes domains $d \in$ \{art, clipart, product, real\}, with 15,588 samples and 65 classes. 
TerraIncognita~\cite{beery2018recognition} includes photos of wild animals taken by a camera at different locations.
Following~\cite{gulrajani2020search}, we used datasets of $d \in$ \{Location 100, Location 38, Location 43, Location 46\}, with a total of 24,788 samples and classes.
We show random samples from each dataset.

VLCS includes four photo datasets, so many objects unrelated to class are captured together. 
We conjecture that training in the source domain can help the model capture the correspondence between images and labels.

However, from the PACS dataset, we can find that the object corresponding to each image is straightforward and clear. 
This would be a relatively simple task for a CLIP trained on large-scale data. 
However, the shift of each Domain is evident, and if the large model is trained on the source domain, it will lead to performance degradation.

\begin{figure}[h]
\begin{center}
\includegraphics[width=1.0\linewidth]{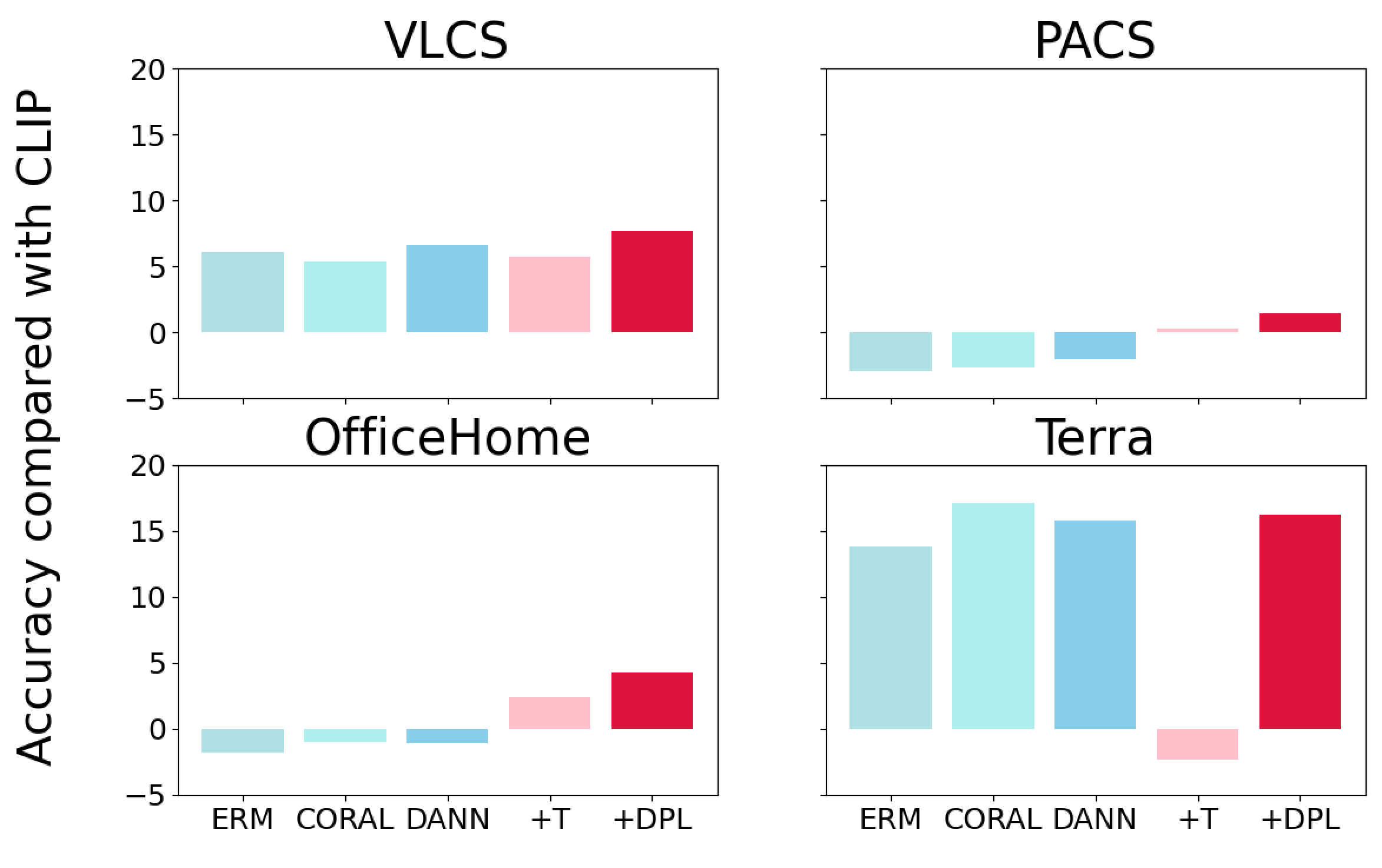}
\end{center}
  \caption{
  The visualization of the results compared with CLIP.
  The methods fine-tuning backbones are colored in blue, and freezing backbones in red.
  +T indicates using template prompt for CLIP.
}
\label{fig:detail_result}
\end{figure}

\begin{figure}[h]
\begin{center}
\includegraphics[width=1.0\linewidth]{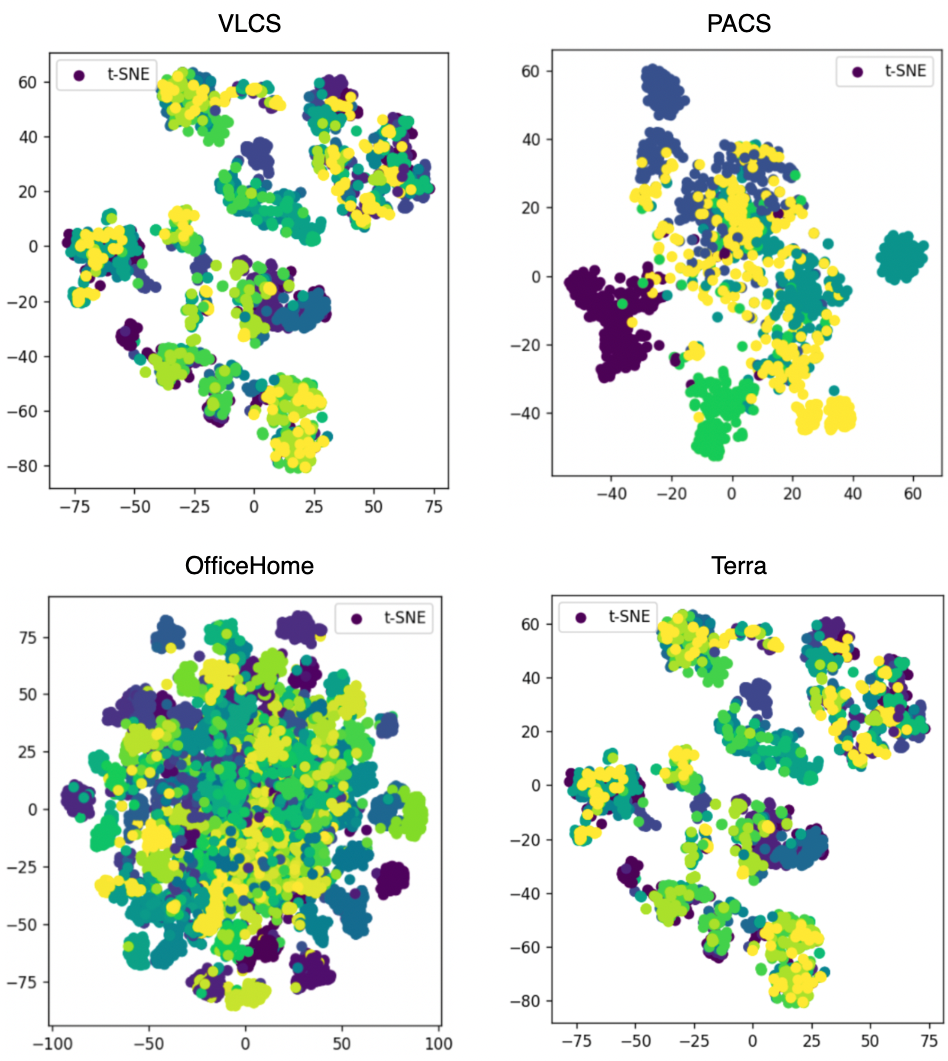}
\end{center}
  \caption{
  The t-SNE visualization of zero-shot CLIP on each dataset.
  We sampled 100 images for each class randomly. If the number of the class is less than 100, we sampled all of them.
  We used default numbers for all parameter of t-SNE, like number of componentsint is 2.
}
\label{fig:t-sne}
\end{figure}

The dataset characteristics of OfficeHome resemble those of PACS in general, which explains our experimental results\autoref{fig:detail_result} that fine-tuning hurts CLIP performance on PACS and OfficeHome.
Through the t-SNE visualization~\footnote{https://scikit-learn.org/stable/modules/generated/sklearn.manifold.TSNE.html} of zero-shot CLIP's embeddings\autoref{fig:t-sne}, we can find that zero-shot CLIP has a specific feature separation on PACS and OfficeHome. In contrast, VLCS and Terra have various color classes overlapping each other, which need to be trained.
We hope this additional analysis can give a better understanding of our results.

Finally, we found that the models perform well if fine-tuning on Terra.
This is because there is no way for Terra's image-label correspondence to be learned during pre-training.
It is worth noting that the SoTA method MIRO~\cite{cha2022domain} not only uses a more advanced backbone than CLIP but also fine-tunes the backbone and adds the SWAD technique. 
These factors lead to the fantastic result of MIRO reaching 64.3\% on Terra.

The real-world domain generalization is similar to the case of Terra~\cite{koh2021wilds}. Therefore, we believe that similar to the MIRO, and it is essential to study FM in the DG domain.

\begin{figure*}[h]
\begin{center}
\includegraphics[width=1.0\linewidth]{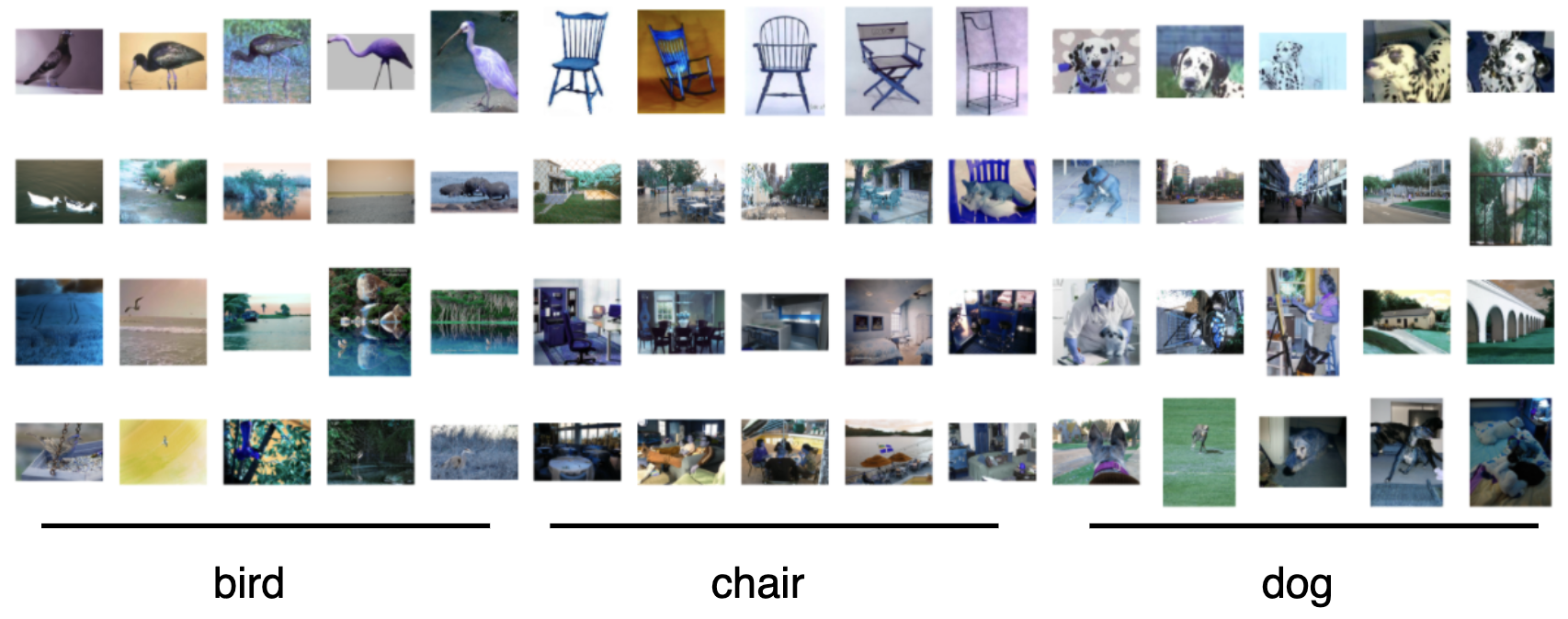}
\end{center}
  \caption{
  The image examples in VLCS.
}
\label{fig:vlcs}
\end{figure*}

\begin{figure*}[h]
\begin{center}
\includegraphics[width=1.0\linewidth]{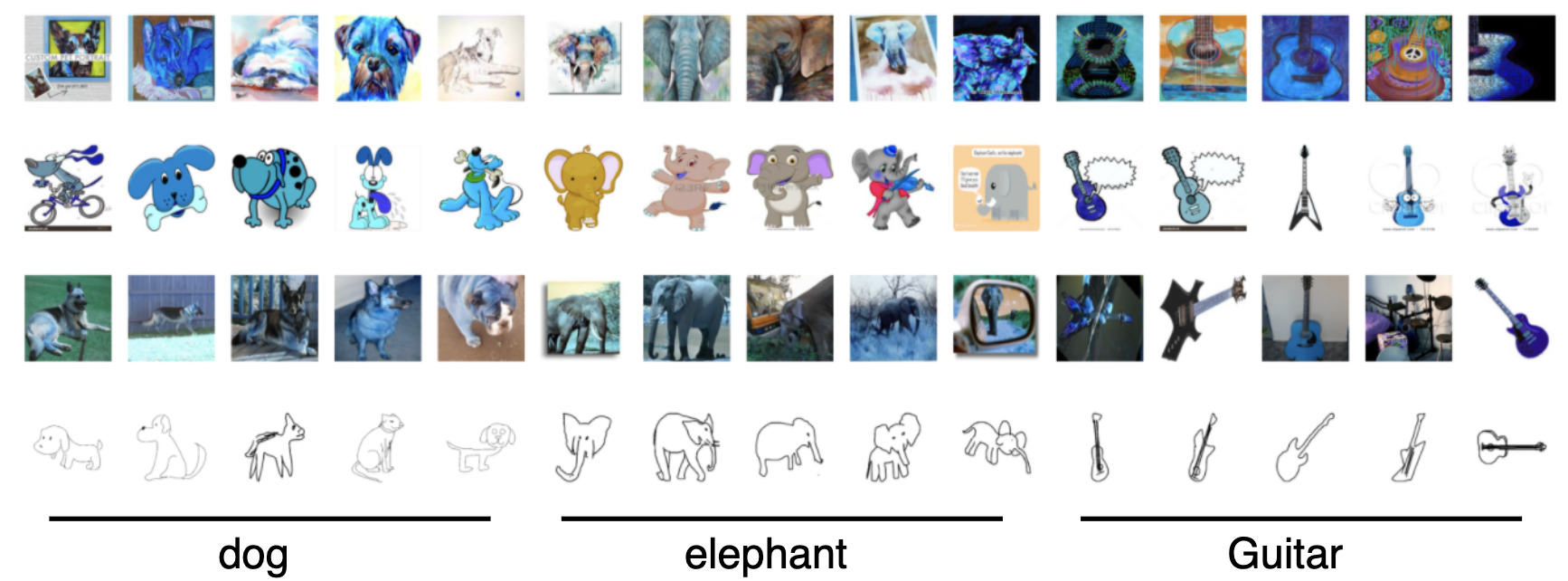}
\end{center}
  \caption{
  The image examples in PACS.
  From the first row to the fourth row are `art painting', `cartoons', `photos', and `sketches'.
}
\end{figure*}

\begin{figure*}[h]
\begin{center}
\includegraphics[width=1.0\linewidth]{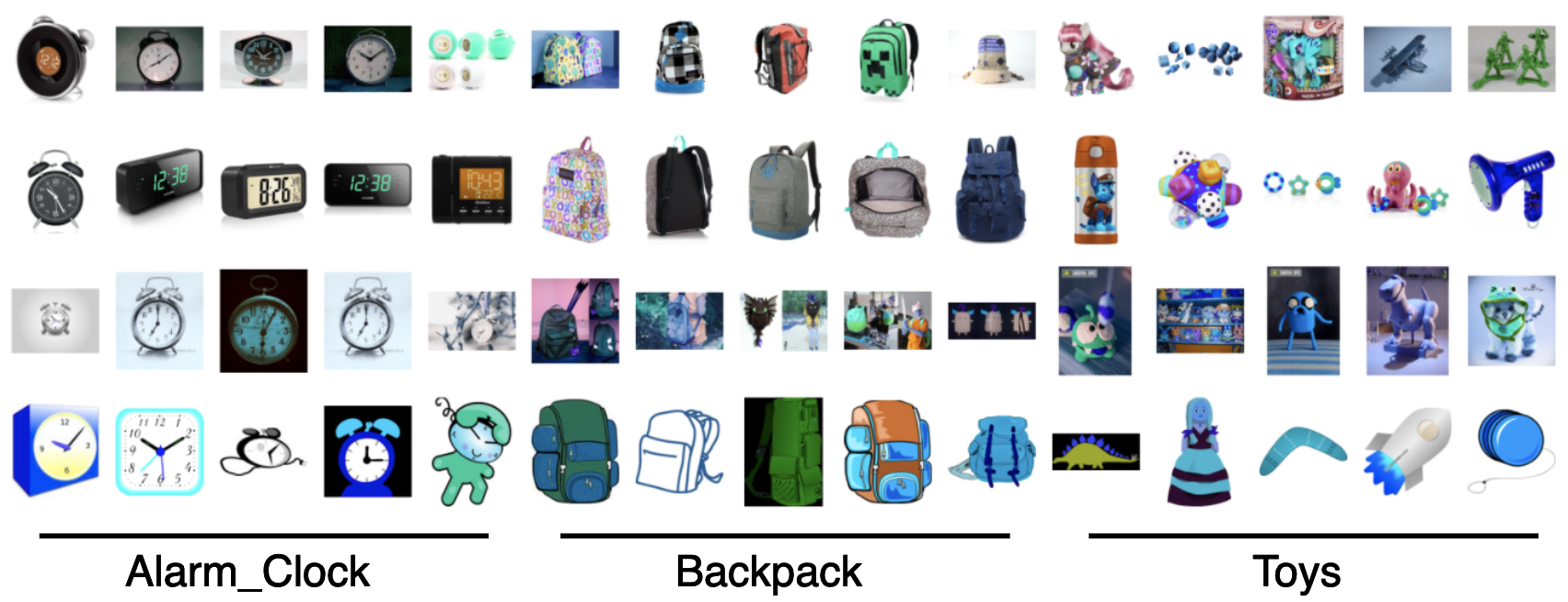}
\end{center}
  \caption{
  The image examples in OfficeHome.
  From the first row to the fourth row are `art', `product', `real', and `clipart'
}
\label{fig:office_home}
\end{figure*}

\begin{figure*}[h]
\begin{center}
\includegraphics[width=1.0\linewidth]{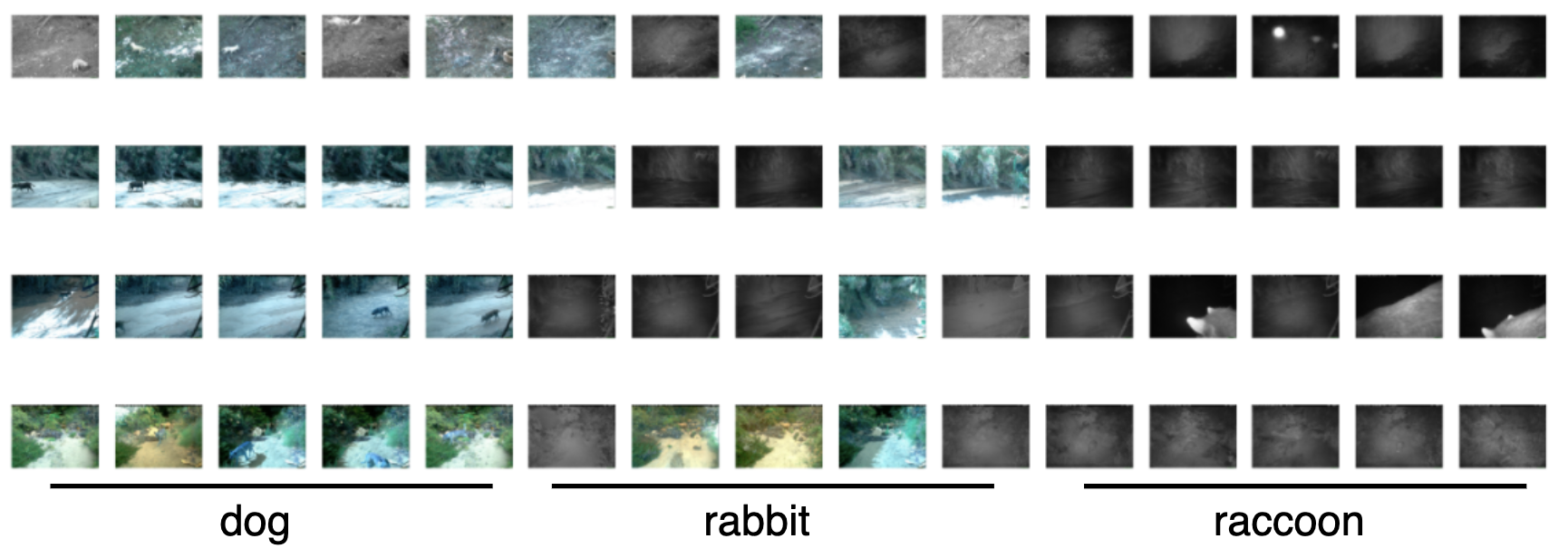}
\end{center}
  \caption{
  The image examples in TerraIncognita.
  From the first row to the fourth row are different location domains.
  From the dataset we can conjecture that Terra is a difficult dataset to benefit from pre-training backbone. This is because his domain is very specific and difficult to classify.
}
\label{fig:terra_incognita}
\end{figure*}

\end{document}